\documentclass[lettersize,journal]{IEEEtran}
\UseRawInputEncoding    
\IEEEoverridecommandlockouts                              %
\usepackage{lineno}
\usepackage[T1]{fontenc} 
\usepackage{enumerate}
\usepackage{amsfonts}
\usepackage{svg}
\usepackage{subcaption}
\usepackage{diagbox}
\usepackage{wrapfig}
\usepackage{multirow}
\usepackage{soul}
\usepackage[colorinlistoftodos]{todonotes}
\usepackage[font=small,skip=2pt]{caption}
\usepackage{float}
\usepackage{subcaption}
\usepackage[ruled,vlined]{algorithm2e}
\makeatletter
\let\NAT@parse\undefined
\makeatother
\usepackage{hyperref}
\usepackage{eqnarray}
\usepackage{array}
\usepackage{verbatim}
\usepackage{amsmath}
\usepackage{textcomp}
\usepackage{stfloats}
\usepackage{balance}
\usepackage{graphicx}
\title{\LARGE \bf Learning Generalizable Vision-Tactile Robotic Grasping Strategy for Deformable Objects via Transformer
}
\author{
Yunhai Han$^{1}$, Kelin Yu$^{1}$, Rahul Batra$^{1}$, Nathan Boyd$^{1}$, Chaitanya Mehta$^{1}$, Tuo Zhao$^{2}$, Yu She$^{3}$, Seth Hutchinson$^{1}$, and Ye Zhao$^{1, *}$
\vspace{-0.6cm}
\thanks{$^{1}$Institute for Robotics and Intelligent Machines, Georgia Institute of Technology, Atlanta, GA 30332, USA (email: \{yhan389, kyu85, gtg693m, nboyd31, cmehta43, seth, yezhao\}@gatech.edu).
}
\thanks{$^{2}$School of Industrial and Systems Engineering, Georgia Institute of Technology, Atlanta, GA 30332, USA (email: tourzhao@gatech.edu).
}
\thanks{$^{3}$School of Industrial Engineering, Purdue University, 610 Purdue Mall, West Lafayette, IN 47907, USA (email:  shey@purdue.edu).}
\thanks{$*$ Corresponding author}
\thanks{Special credit to GelSight hardware support from E. Adelson's lab at MIT.}  
}%
\begin{document}
\markboth{IEEE/ASME TRANSACTIONS ON MECHATRONICS}%
{How to Use the IEEEtran \LaTeX \ Templates}
\maketitle
\begin{abstract}
Reliable robotic grasping, especially with deformable objects such as fruits, remains a challenging task due to underactuated contact interactions with a gripper, unknown object dynamics and geometries. 
In this study, we propose a Transformer-based robotic grasping framework for rigid grippers that leverage tactile and visual information for safe object grasping. 
Specifically, the Transformer models learn physical feature embeddings with sensor feedback through performing two pre-defined explorative actions (pinching and sliding) and predict a grasping outcome through a multilayer perceptron (MLP) with a given grasping strength. 
Using these predictions, the gripper predicts a safe grasping strength via inference. Compared with convolutional recurrent networks, the Transformer models can capture the long-term dependencies across the image sequences and process spatial-temporal features simultaneously. 
We first benchmark the Transformer models on a public dataset for slip detection. Following that, we show that the Transformer models outperform a CNN+LSTM model in terms of grasping accuracy and computational efficiency. We also collect a new fruit grasping dataset and conduct online grasping experiments using the proposed framework for both seen and unseen fruits. {In addition, we extend our model to objects with different shapes and demonstrate the effectiveness of our pre-trained model trained on our large-scale fruit dataset.} Our codes and dataset are public on GitHub \footnote{\href{https://github.com/GTLIDAR/DeformableObjectsGrasping.git}{https://github.com/GTLIDAR/DeformableObjectsGrasping.git}}.
\end{abstract}
\vspace{-0.4cm}
\section{INTRODUCTION}
\IEEEPARstart{R}{obot} manipulation has been widely used in industries for decades, but mostly for repetitive tasks in structured environment where there is little uncertainty or contact deformation in manipulated objects.
For the tasks where object contact parameters are prone to vary, such as fruit grasping, they are still challenging for robotic systems \cite{pickplace2011}. Loose grips with small grasping forces can cause objects to slip, while large grasping forces can cause damage. Additionally, object contact geometry and frictional properties may also affect the optimal grasping forces for safe grasping. 
To learn general-purpose grasping skills, robots need to leverage 
with dense notions of contact information from in-hand interactions. 
\begin{figure}[htb!]
\centering
\includegraphics[width=1\linewidth]{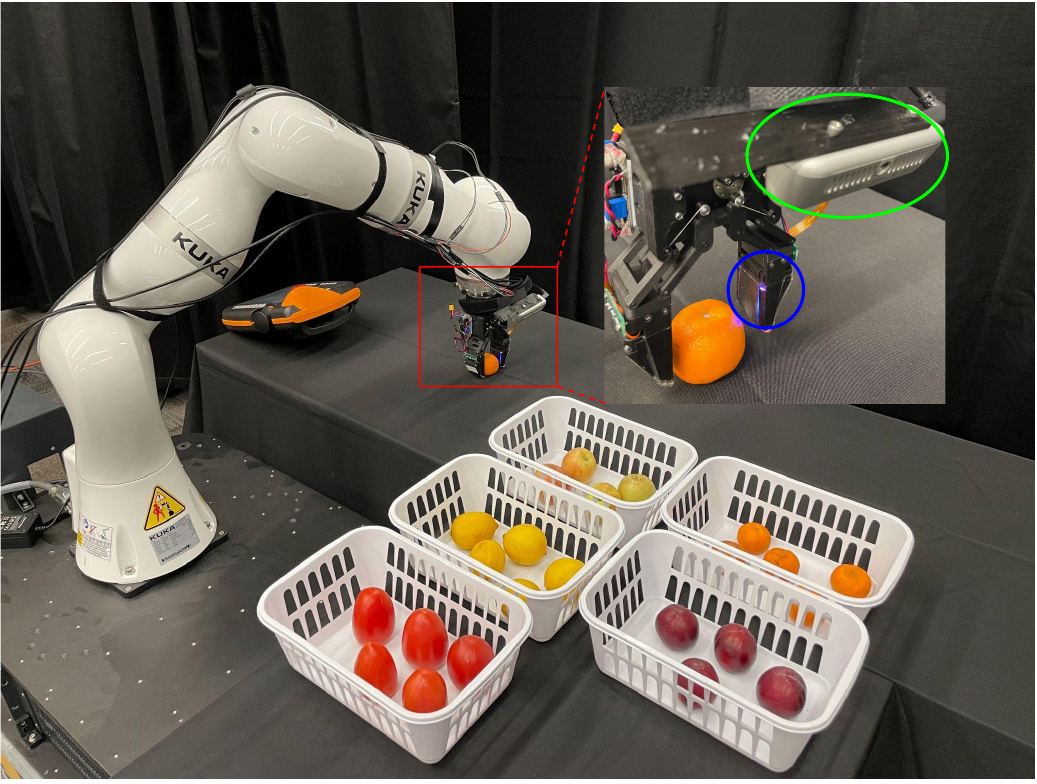}
\caption{Demonstration of our robotic grasping experimental set-up. The robot gripper safely grasp the fruits on the table and sorts them into the target bins via the learned framework.
Robot setup: a KUKA LBR iiwa robot is equipped with a gripper (red box), of which both fingers are equipped with a GelSight sensor (blue circle). A Realsense D$435$ is mounted above the gripper (green ellipsoid).
}
  \label{fig:cover}
  \vspace{-0.5cm}
\end{figure}



To model the dynamic interactions between the object and its environment, vision-based sensing frameworks have been studied based on a sequence of visual observations obtained by external cameras \cite{xu2019densephysnet}, \cite{jiajun2015Galileo}.
However, these methods are not sensitive to the dense local deformation near contact regions, which could lead to errors between the perceived and actual states of a grasp.
To address this issue, 
tactile sensing has gained increasing popularity recently \cite{Luo2017tactileReview}. Among various tactile sensors, the ones with internal cameras, such as GelSight sensor \cite{wang2021gelsight}, have the capability of capturing high-resolution image data regarding local contact geometry. Other tactile designs \cite{Yuan2017gelsight}, \cite{Punyo2020} have also demonstrated a variety of manipulation tasks with similar methods. Compared to force sensors, tactile sensors can capture an object's deformation during contact. Moreover, tactile data can be readily integrated by modern learning methods for classification and task-oriented control policy learning \cite{wang2021swingbot}. 
 In \cite{li2018slip}, \cite{calandra2017feeling}, they demonstrated that grasping performance can be significantly improved by incorporating visual and tactile sensing.

In this paper, we employ two state-of-the-art Transformer models -- TimeSformer \cite{gberta_2021_ICML} and ViViT \cite{arnab2021vivit} -- to 
determine safe grasping forces
from the visual and tactile image sequences collected during pre-designed explorative actions (e.g., pinching and sliding). 
The idea of designing task-oriented explorative actions is inspired by \cite{wang2021swingbot} and the motivations of introducing the Transformer models are: 1) compared with recurrent networks, such as LSTM, they do not suffer from the forgetting issue, 2) compared with convolutional networks used for extracting local features, they have larger receptive fields that are helpful to understand the global context, and 3) compared with CNN+LSTM models for processing image sequences, they can extract the spatial-temporal features simultaneously. While for CNN + LSTM models, the per-frame spatial features are always encoded (CNN) prior to the temporal decoding (LSTM). Thus, the Transformer models are more adaptable to complex tasks.
In our framework, the Transformer models learn low-dimensional embeddings in a supervised fashion for each sensor modality and then output a fused physical feature embedding. Firstly, we take this embedding as input and combine it with a given grasping force threshold to predict the final grasping outcomes through a multilayer perceptron (MLP). The grasping outcomes are categorized into three labels: safe grasping, slippery, and potential damage. A force threshold for safe grasping is then searched for using the learned predictor during online deployments.
Secondly, the fused physical feature embedding is used to classify grasped fruit types through a different MLP layer in order to place them into separate bins automatically. 

To begin with, we benchmark the Transformer models against a CNN+LSTM model on a public dataset for slip detection \cite{li2018slip}. 
Both Transformer methods (TimeSformer and ViViT) outperformed the CNN+LSTM model by $3.1\%$ and $2.0\%$ in detecting slip, and are much computationally efficient, making them more suitable for online tasks. Then,
in order to validate the grasping framework, we perform grasping experiments on various deformable fruits for data collection. We train the models using both camera and GelSight inputs and test their performance via grasping outcome classifications on unseen fruits and online grasping success rate for both seen and unseen fruits. {We additionally conduct sensitivity analysis for visual and tactile images to evaluate the framework's robustness to images with varying qualities. Furthermore, we employ our pre-trained model, the one trained on a dataset of multiple fruits, for training with an unseen banana dataset. Notably, employing the pre-trained model accelerates the training process by four times and leads to a $10$\% improvement in training accuracy in the new banana dataset.}

The contributions of our work are summarized as follows: 
\vspace{-0.2cm}
\begin{itemize}
\item We propose a Transformer-based grasping framework for fruit grasping and demonstrate the superior efficacy and efficiency of the Transformer models against a CNN+LSTM baseline model.
\item We design a learning-based control framework that incorporates safe grasping force estimation using tactile \& visual information obtained via two explorative actions: pinching and sliding, which do not require any prior knowledge of physical contact or geometrical models. Besides, the control parameter is directly formulated as the depth value read from the tactile feedback without any aid of external force-torque sensors.
\item We experimentally evaluate the proposed grasping framework on a diverse set of fruits and achieve an end-to-end demonstration of fruit grasping. Besides, by performing the attention analysis, we show that the trained Transformer models take advantage of the attention mechanism to: i) incorporate more contact area information for the grasping task, such as local contact region in tactile images and fruit surface near the gripper's fingertips in visual images; and to ii) capture long-term dependencies between initial and final grasping status. {Furthermore, we conduct detailed sensitivity analysis for tactile images and visual images with different data qualities. We evaluate the robustness of our transformer models according to the difference in outputs.}
\item {By employing our model trained on a large fruit dataset as the pre-trained model for our collected banana dataset, we observe improved training efficiency and higher success rates in both the training process and online grasping experiments.}
\end{itemize}
\vspace{-0.5cm}
\section{Related Work} \label{section: relatedwork}
\vspace{-0.1cm}
\subsection{Robotic Grasping}  
\vspace{-0.1cm}
Robotic grasping has been a widely explored topic using numerous gripper designs and sensor modalities \cite{gripreview2016}. 
The sensorized hand proposed in \cite{friedl2020clash} uses a multi-model observer framework and a variety of sensors, including cameras, tendon force sensors, and proximity sensors, to achieve successful grasping.
In \cite{pozzi2022grasping}, the presented work proposes a vision-based framework for grasp learning of deformable objects using an anthropomorphic, under-actuated, compliant hand. It renders a promising future direction to employ advanced hand mechanisms for grasping. Recently, motivated by human's intense dependence of tactile feedback for the grasping process, tactile sensors have thus begun to play an important role in robotic grasping \cite{jiang2022shall}. In \cite{Calandra2018Regrasp}, the authors used deep-learning methods to obtain a grasping policy for rigid grippers. However, the grasping success rate on deformable objects was not ideal since they only adjusted the grasping position but fixed the grasping force. 
The work of  \cite{Dongeon2020OptimalGrasping} estimated the optimal grasping force empirically but assumed the object weights were known.
In \cite{Yazhan2020TowardsLearning} \cite{Siyuan2019Slip}, the gripper's opening was controlled to stabilize the grasped objects under external disturbances by detecting the slip occurrences. Similarly, the work in \cite{wettels2009grip} made use of tactile sensing to stabilize the grasped object by controlling the grasping force.  However, all of these studies assumed that the objects were already steadily grasped in hand. In this work, we aim at estimating safe grasping force for deformable objects through a learning framework.
\vspace{-0.45cm}
\subsection{Vision-tactile Sensor Fusion}
\vspace{-0.1cm}
We can improve the manipulation performance by fusing the information obtained from visual and tactile sensors. In \cite{calandra2017feeling}, they proposed a multi-modal sensing framework for grasping outcome prediction. Their subsequent work in \cite{Calandra2018Regrasp} investigated a learned regrasp policy based on visuo-tactile data after executing an initial grasp. Their results indicated that incorporating tactile readings substantially improves grasping performance. However, the manipulated objects used in their experiments are primarily rigid objects, which do not require accurate force control. 
In other works \cite{li2018slip, Justin2019IdenObject, Yuan2017CVPR}, they used CNN + LSTM models to classify the slip occurrence, to recognize the object instance, and to perceive the physical properties of objects. Nonetheless, these methods can only be used for classification tasks and are not applicable to learning control policy for safe manipulation.
\vspace{-0.45cm}
\subsection{Transformers for Robotics}
\vspace{-0.1cm}
Transformer models were originally proposed for natural language processing (NLP) \cite{Vaswani2017Attention} and computer vision (CV)  \cite{dosovitskiy2021image}\cite{gberta_2021_ICML}\cite{arnab2021vivit}\cite{xu2022v2x}. Recently, Transformers have drawn increasing attention in robotics. The authors of  \cite{shridhar2022perceiver} proposed a Transformer framework for tabletop tasks, which encodes language goals and RGB-D voxel observations and output discretized $6$-DoF actions. In \cite{Monastirsky2022TransformerThrow}, they explored the use
of Transformers to predict robot action commands for accurate object throwing. The study of \cite{yang2022learning} addressed quadrupedal locomotion tasks using Reinforcement Learning (RL) with a Transformer-based model. All of these works showed significant improvements over baseline methods on task performance and training efficiency. However, to the best of our knowledge, no existing study has ever explored the use of  Transformers for robotic grasping using tactile and visual images.


\vspace{-0.45cm}
\section{Methods} \label{section: methodology}
\vspace{-0.05cm}
In this section, we describe the details of the grasping framework and each Transformer model. In order to give robots the ability to estimate the safe grasping force, we first let the robot obtain physical information about the target objects (fruits in this work) by performing
two explorative actions, \textit{pinching} and \textit{sliding}, on the objects. To avoid any potential damage, these actions have minimum interaction with the objects. To monitor the interactions and record the data, the robot is equipped with two different sensors. Next, a force threshold for safe grasping will be searched for via inference using the obtained physical information and adopted for execution. In the following, we first describe the sensors in Section \ref{sec:sensor}, and then we discuss the Transformer models in Section \ref{sec:Transformer}, and finally we propose the grasping framework in Section \ref{sec:framework} and Section \ref{sec:fruit_classi}.
\vspace{-0.5cm}
\subsection{Sensing Modalities} \label{sec:sensor}
\vspace{-0.05cm}
\subsubsection{Tactile} The GelSight \cite{wang2021gelsight} sensor provides the robot with dense visual information (high-resolution image) about the contact region between the objects and the robot's fingertips. For this purpose, the contact surface of the sensor is covered with a soft elastomer such that the sensor can measure the object's compliance by observing the elastomer's vertical and lateral deformation. In out experiments, the gripper has two GelSight sensors installed on the fingertips, but we only use one to demonstrate a minimum system setup.
\subsubsection{Vision} A RealSense D$435$ camera is used in this work, and we only consider the RGB datastream. The camera is wrist mounted at an angle of $15$ degrees such that the image is centered on grasped objects (see Fig.~\ref{fig:cover} for the setup).
\vspace{-0.45cm}
\subsection{Transformer Model} \label{sec:Transformer}
\vspace{-0.05cm}
We apply the Transformer models for two robotic manipulation tasks: slip detection and safe grasping force estimation. For slip detection, we replace the CNN + LSTM model used in \cite{li2018slip} with the Transformer models but keep the last Fully Connected (FC) layer with two outputs (i.e., a stable grasp or slip) as the final classification results. For safe grasping force estimation, the outputs from Transformer models are used as inputs to the subsequent models in the grasping framework that will be thoroughly described in Section~\ref{sec:framework} \& Section~\ref{sec:fruit_classi}. 

Two lightweight Transformer models are explored for robotic tasks in this work: TimeSformer \cite{gberta_2021_ICML}, ViViT \cite{arnab2021vivit}. Each model uses similar self-attention mechanism, 
which brings main advantages over CNN+LSTM models, while the difference of these models lies in the factorizing strategy for the spatial-temporal attention. 
\begin{figure}[t] 
\centering
\includegraphics[scale=0.42]{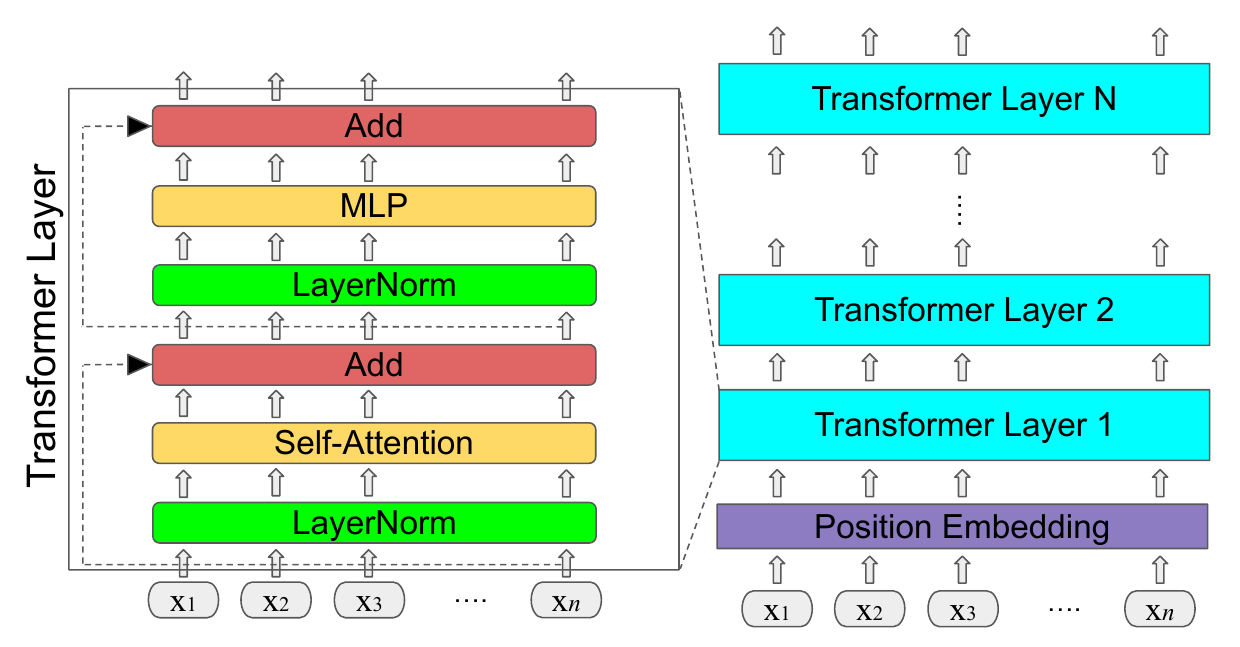}
\caption{An illustration of the Transformer model structure. The left figure shows one Transformer layer and the right figure shows the encoding structure. $\mathrm{x}_1, \mathrm{x}_2, \mathrm{x}_3, \ldots, \mathrm{x}_n$ are the input vectors which are first linearly embedded and then added with the position embeddings before Transformer layer 1.} 
\label{fig:Layer}
\vspace{-0.5cm}
\end{figure}
A Transformer layer contains a self-attention layer 
and a MLP layer. To stack the Transformer layers for a deeper encoding structure, the MLP layer does not change the vector size. 
Also, before and after both layers, there is a LayerNorm and a residual connection, respectively. One Transformer layer is shown in Fig.~\ref{fig:Layer} (left-side subfigure), where the outputs of the current layer will be the inputs for the next (right-side subfigure). Before the first Transformer layer, all the input vectors will be linearly embedded and then added with position embeddings, the elements of which represent the positions of each vector, to retain the useful sequence knowledge \cite{Vaswani2017Attention}.

{\textbf{Self-Attention Mechanism}
The self-attention mechanism allows all the inputs to interact with each other and identify the ones that should be paid more attention to, which renders their main advantages over CNN+LSTM models.
Specifically, this mechanism can be described as mapping a query ($Q$) and a key ($K$)-value ($V$) pair to the outputs.}

{For a single self-attention block (or a single head), the query, key, and value vectors can be computed by projecting the same input matrix $\mathrm{X}\in\mathbb{R}^{n \times d_x}$ (each row of $\mathrm{X}$ corresponds to an input vector with size $d_x$) to $\mathrm{Q}, \mathrm{K}, \mathrm{V}$ as follows:
\vspace{-0.15cm}
\begin{equation}\label{eqn: QKV}
\mathrm{Q} = \mathrm{X}\mathrm{W}^{Q}, \mathrm{K} = \mathrm{X}\mathrm{W}^{K}, 
\vspace{-0.15cm}\mathrm{V} = \mathrm{X}\mathrm{W}^{V}
\end{equation}
where $\mathrm{W}^{Q} \in \mathbb{R}^{d_x \times d_k}$, $\mathrm{W}^{K} \in \mathbb{R}^{d_x \times d_k}$ and $\mathrm{W}^{V} \in \mathbb{R}^{d_x \times d_v}$ are learnable matrices with $d_k=d_v=d_x$.}

{The outputs of the self-attention mechanism are obtained through Eqn.~\ref{eqn:attention}, which represents the weighted sums of the value vectors ($\mathrm{V}$) with the weights assigned based on a compatibility function between the query vectors ($\mathrm{Q}$) and the corresponding key vectors ($\mathrm{K}$) at the same vector index. Note that, the dot-product between $\mathrm{Q}$ and $\mathrm{K}$ is scaled by $\sqrt{d_{k}}$ as suggested in \cite{Vaswani2017Attention}.
\vspace{-0.25cm}
\begin{equation}\label{eqn:attention}
\text {Attention}(\mathrm{Q}, \mathrm{K}, \mathrm{V}) =\text{softmax} \left(\frac{\mathrm{Q} \mathrm{K}^{T}}{\sqrt{d_{k}}}\right) \mathrm{V}
\end{equation}
\vspace{-0.25cm}
\begin{equation}\label{eqn:attention2}
\text {SingleHead}(\mathrm{X})=\text{Attention}
\vspace{-0.1cm}(\mathrm{Q}, \mathrm{K}, \mathrm{V}) \mathrm{W}^{O} 
\end{equation}
As shown in Eqn.~\ref{eqn:attention2}, another learnable matrix $\mathrm{W}^{O} \in \mathbb{R}^{d_v \times d_x}$ projects the intermediate results to a new matrix with the same dimension as $\mathrm{X}$.}
        
{
In practice, to allow the model to attend to information from different combinations of input space representations, we employ a MultiHead strategy by projecting the $\mathrm{Q}, \mathrm{K}, \mathrm{V}$ matrices $h$ times with different sets of weights ${W^{Q}_i, W^{K}_i, W^{V}_i}$ for $i = 1, \ldots h$. This strategy leads to more effective representations and improved performance. As a result, it is always employed by Transformer models.}

{In addition to the self-attention blocks, there is a fully-connected 
MLP layer applied to each vector position separately and identically. It has two linear transformations and a GeLU activation function in between.}

\textbf{Factorization of Spatial-Temporal Attention:} For image-based tasks, to generate input vectors from raw image(s), in \cite{dosovitskiy2021image}, they split an image into fixed-size patches and embed each of them via linear transformation. 
Our framework handles image sequences instead of single images and must consider the temporal dimension within each self-attention layer. To accomplish this, we incorporate spatial-temporal factorization using TimeSformer \cite{gberta_2021_ICML} and ViViT \cite{arnab2021vivit}.

\begin{figure}[t]
\centering
\includegraphics[scale=0.44]{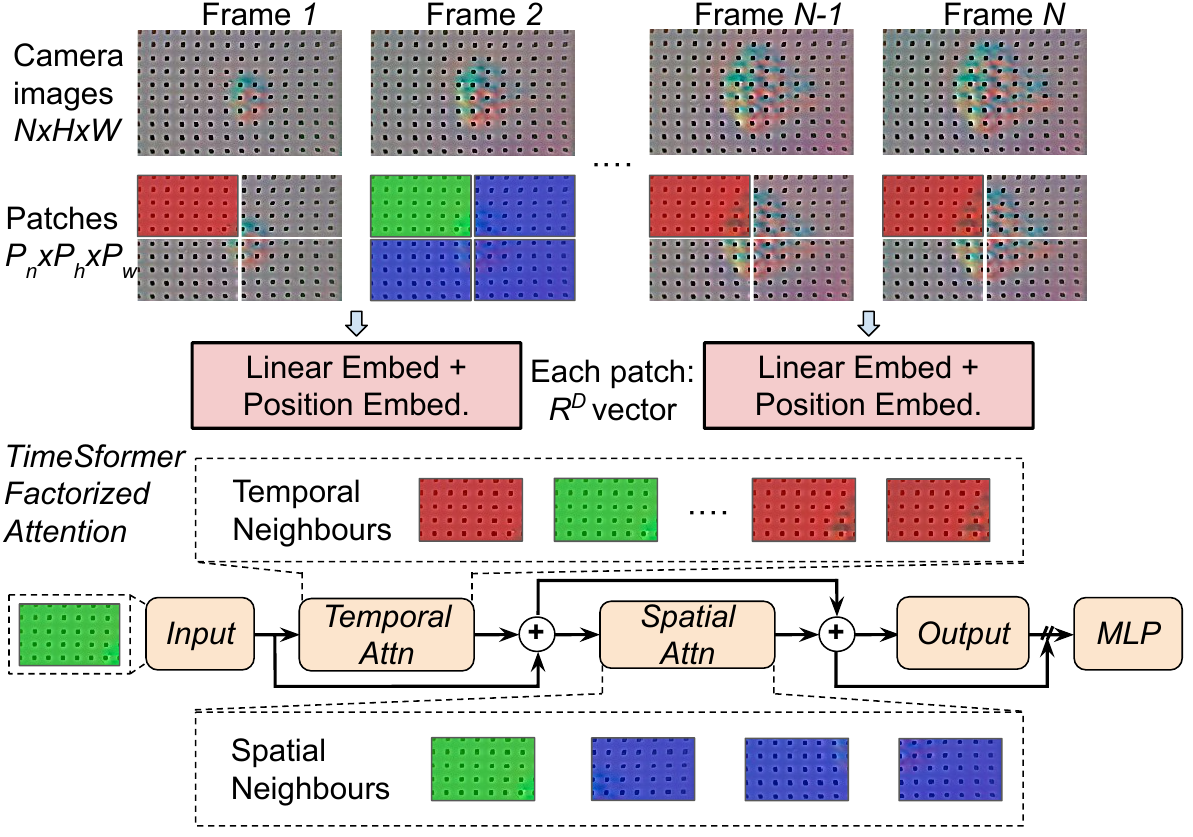}
\caption{Visualization of space-time attention approach for TimeSformer. The top three rows show an input GelSight image sequence, the generated $4N$ image patches and the patch embeddings. We denote one image patch in green and its spatial-temporal neighbours 
in blue and red, respectively. 
Within each self-attention layer, for each image patch, the attentions across temporal neighbours and spatial neighbours are sequentially processed and the output will be the input of the MLP layer in Transformer model after LayerNorm.}
\label{fig:TransformerArchitecture}
\vspace{-0.5cm}
\end{figure}

\textbf{TimeSformer:}
In this model, spatial-temporal dimensions are processed sequentially: within each self-attention layer, the attention is first applied on the temporal dimension of the inputs at the same spatial position, followed by the spatial dimension among all inputs from the same temporal position. There are also residual connections between each operation. This approach is visualized in Fig.~\ref{fig:TransformerArchitecture}.
In our work, the input image sequence is denoted as $\mathrm{X}_I \in \mathbb{R}^{N \times H \times W}$, where $N, H, W$ are the number of images, image height pixels, and image width pixels, respectively. We first extract the patches $\mathrm{X}_P \in \mathbb{R}^{P_n \times P_h \times P_w}$, where $(P_h, P_w)$ is the resolution of each patch and $P_n = \frac{NHW}{P_hP_w}$. Next, these patches are flattened and then linearly embedded to vectors of size $D$ with a positional embedding being added to each of them. We further add a CLS (classifier) token to the sequence of embedded vectors, which is designed to extract task-level representations \cite{devlin2019bert} by attending to all the other vectors and {forming an augmented sequence ${X} = \{CLS, x_1, x_2, ..., x_N\}$, where $x_i \in \mathrm{X}_P$.}
{$E({X})$, which is the Transformer encoder function in ViT, then takes ${X}$ as input and generates an encoded sequence of representations $\{h_{CLS}, h_1, h_2, ..., h_N\}$, where $h_i$ is the representation of the $i$-th patch.}
Finally, the output of the CLS token {$h_{CLS}$} 
is used for different tasks. In slip detection (Section.~\ref{sec:slipDec}), it is passed through a MLP layer to classify whether or not a slip occurs. 

\textbf{ViViT:}  
Our implementation of ViViT is similar to TimeSformer, except for the following differences: First, both dimensions are processed in parallel. Specifically,
half of the heads attend to the spatial dimension and the other half to the temporal dimension (factorized dot-product attention). We then combine each output by concatenation and add a linear transformation to halve the size. Second, there is no CLS tokens added to the embedded input vectors because of the ambiguities when dot-producting the temporal and spatial attention. Instead, we take the average of all patch outputs from the last Transformer layer and pass it (size $D$) to the MLP layer to classify whether or not a slip occurs
\vspace{-0.55cm}
\subsection{Grasping Framework for Safe Force Estimation} \label{sec:framework}
\vspace{-0.1cm}
The main goal of our grasping framework is to predict the grasping outcome given a grasping force threshold and to estimate the force threshold for safe grasping via inference.
\subsubsection{Grasping Outcome Prediction} 
\begin{figure*}[htb!]
\centering
\includegraphics[width=0.9\textwidth]{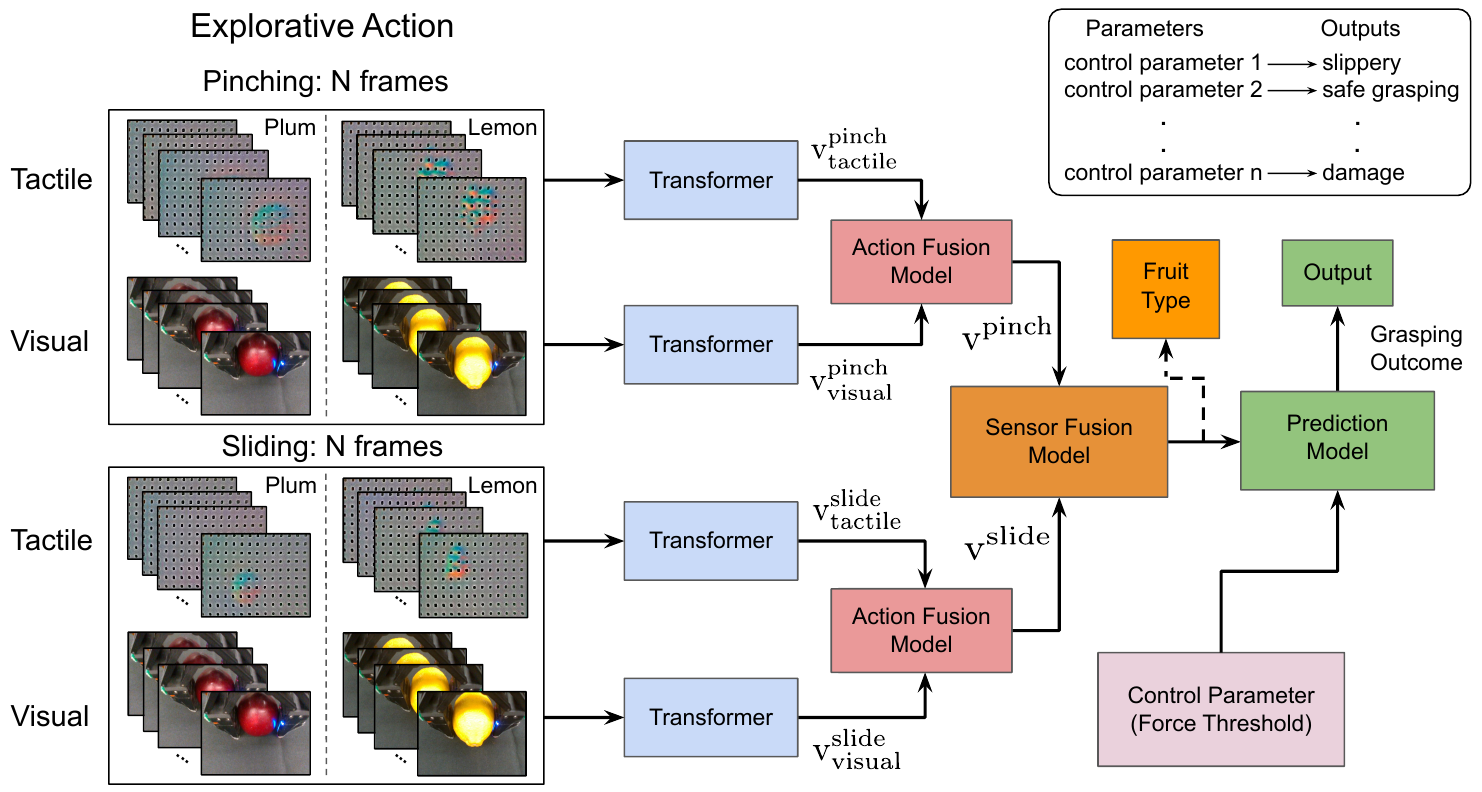}
\caption{\textbf{Overview of Grasping Framework.} The robot first performs two explorative actions: 1). pinching the fruit, 2). sliding  along the fruit surface in the optical axis. Many fruit examples can be found on our GitHub page. Each image sequence is processed by an individual Transform network into a vector of size $D$. The fusion models concatenate these vectors and project it into a low-dimensional fused physical feature embedding. This embedding is further processed by a Fruit Classification model to classify the grasped fruit type. Besides, the prediction model takes the same embedding and control parameter (force threshold) as inputs and predicts the final grasping outcome. Through inference, a set of control parameters is first generated and then the parameter with the safe grasping outcome is selected to perform online grasping. This procedure is shown in the top-right black box. If there are multiple viable choices, we select the \textbf{average} value. }
\label{fig:framework}
\vspace{-0.4cm}
\end{figure*}

As shown in Fig.~\ref{fig:framework}, this framework is composed of five main components: Control Parameter (Force Threshold), Transformer, Sensor Fusion model, Action Fusion model, and Prediction model. 

\textbf{Force Threshold:}\label{sec:normal_force} GelSight is a vision-based tactile sensor, which lacks the capability of estimating the grasping force (contact normal force) directly. To address this issue, the authors in \cite{Yuan2017gelsight} showed that the contact normal force can be estimated from the depth value (unit: pixel) with accurate gel calibration. On the other side, the work in \cite{Yu2021Wire} directly used the mean value of the marker displacement provided by GelSight images to approximate the resultant frictional force.
Inspired by this work, we employ the maximum depth value as the approximation of grasping forces. 
If the maximum depth value feedback is larger than the selected threshold for three continuous frames when running the framework, the gripper will begin to grasp the fruit. Also, the force threshold will be sent into the Prediction Model. Note that the unit of force threshold is pixel, which will be omitted in Sec. \ref{section: experiments} for readability.

\textbf{Transformer:} For each explorative action, the image sequence from a sensor modality outputs a vector of size $D$ via the Transformer models, as thoroughly described in Sec. \ref{sec:Transformer}. In Fig.~\ref{fig:framework}, since we pre-define two explorative actions and there are two sensor modalities, we have 4 vectors: $\mathrm{v}^{\rm pinch}_{\rm visual}$, $\mathrm{v}^{\rm pinch}_{\rm tactile}$, $\mathrm{v}^{\rm slide}_{\rm visual}$, $\mathrm{v}^{\rm slide}_{\rm tactile}$.

\textbf{Action Fusion model \& Sensor Fusion model:} We concatenate each two vectors obtained from the same exploration action and achieve: $\mathrm{v}^{\rm pinch} = [\mathrm{v}^{\rm pinch}_{\rm visual}, \mathrm{v}^{\rm pinch}_{\rm tactile}]$ and $\mathrm{v}^{\rm slide} = [\mathrm{v}^{\rm slide}_{\rm visual}, \mathrm{v}^{\rm slide}_{\rm tactile}]$. \vspace{0.07cm}{We then fuse them as a vector $\mathrm{v}^{\rm fused} = [\mathrm{v}^{\rm pinch}_{\rm visual}, \mathrm{v}^{\rm pinch}_{\rm tactile}, \mathrm{v}^{\rm slide}_{\rm visual}, \mathrm{v}^{\rm slide}_{\rm tactile}] \in \mathbb{R}^{4\times D}$. Then, we use a linear transformation operation to project it to a low-dimensional space with an output size of $N$.
The linear transformation can be represented as:
\vspace{-0.15cm}
$$\mathbf{\mathrm{Y}^{\rm fused}} = \mathbf{\mathrm{v}^{\rm fused}} 
\vspace{-0.15cm}\cdot \mathbf{W}^\top + \mathbf{b}$$
where $\mathbf{\mathrm{Y}^{\rm fused}} \in \mathbb{R}^{N \times 1}$ is the output vector, which is a fused physical feature embedding.
$\mathbf{W} \in \mathbb{R}^{N \times 4D}$ is the weight matrix which is a learnable parameter in the framework and used to perform the linear mapping, and $\mathbf{b} \in \mathbb{R}^{N \times 1}$ is another learnable bias which is used as an offset to the output.}

\subsubsection{Safe Force Threshold Estimation}
We aim to identify the control parameter, i.e., the safe grasping force threshold. As shown in Fig.~\ref{fig:framework}, the Prediction Model takes the low-dimensional physical embedding obtained from performing two explorative actions and a force threshold candidate as inputs and outputs the upcoming grasping outcome via learnable neural network layers. Next, we can uniformly sample the thresholds and feed each of them into the prediction model and select the one that predicts a safe grasping. When there are multiple viable choices, we select the average value.
\vspace{-0.5cm}
\subsection{Grasping Framework for Fruit Classification}
\vspace{-0.1cm}
\label{sec:fruit_classi}
Our grasping framework includes a goal of fruit type classification for pick-and-place operations. Specifically, we use another MLP network for fruit classfication which takes the fused physical feature embedding of Sensor Fusion model as input and outputs the grasped fruit type (\textit{Fruit Type} Block in Fig.~\ref{fig:framework}). During training, the weights of the Transformer models that generate the embedding are frozen and only the MLP network is trained in a supervised learning scheme. 

\vspace{-0.3cm}
\section{EXPERIMENTS} \label{section: experiments}
In this section, we present our experiments using the Transformer models. The robot setup is shown in Fig.~\ref{fig:cover}.

\vspace{-0.55cm}
\subsection{Transformers for Slip Detection}
\vspace{-0.1cm}
To begin with, we benchmark the Transformer models against a CNN+LSTM model on a public dataset for slip detection with different sensor modalities.
\label{sec:slipDec}
\subsubsection{Experiment Setup}
We conduct experiments for a slip detection task. The dataset released by \cite {li2018slip} is used and can be directly downloaded online \footnote{\href{https://drive.google.com/file/d/1\_iTPxl8TENznXVh-82I26kW9qXrsNsv-/view?usp=share\_link}{https://drive.google.com/file/d/1\_iTPxl8TENznXVh-82I26kW9qXrsNsv-/view?usp=share\_link}}.
During implementation, the entire dataset is split into training, validation, and test sets, where the test data uses unseen objects in the training data. Since the size of dataset is relatively small, we randomly split the dataset five times and train the model on each of them to mitigate the effect of overfitting.
The final detection accuracy on the test set is averaged. For each model, we analyze the performance with three different data source inputs (vision-only, tactile-only, and vision \& tactile). For the CNN+LSTM model, ResNet18 is chosen as the CNN architecture over other options, such as VGG or Inception \cite{li2018slip}, due to its advantage of fewer parameters. As a result, the ResNet18 architecture can be initialized randomly without the need of loading a pre-trained model.
\begin{table}[tb!]
    \begin{tabular}{|c|c|c|c|c|c|}
    \hline
        \multirow{3}{*}{\diagbox[width=9em,height=3\line]{Modality}{Accuracy}{Model}} &  & \multirow{3}{*}{TimeSformer} & \multirow{3}{*}{ViViT} \\ 
            &  CNN + LSTM & & \\
            & ResNet18 & & \\
        \hline
    vision-only    &   71.7\% (0.4\%) & 78.7\% (0.7\%)  & 78.9\% (1.2\%) \\
    \hline
    tactile-only    &  80.6\% (0.8\%) & 81.0\% (0.5\%) & 81.8\% (0.5\%) \\
        \hline
    vision \& tactile &  81.9\% (0.3\%) & \textbf{85.0}\% (0.4\%) & 83.9\% (0.3\%) \\
        \hline \hline
        Execution time of & \multirow{2}{*}{9.61s} & \multirow{2}{*}{2.46s}  & \multirow{2}{*}{2.43s}\\
        feed-forward test & & & \\
        \hline
    \end{tabular}
    \caption{Experimental results on slip detection dataset \cite {li2018slip}. TimeSformer and ViViT outperform the CNN + LSTM method by $3.1$\% and $2.0$\%, respectively. Recorded values are the average across $5$ dataset splits and their variances in parenthesis.}
    \label{tab:slipdetection}
    \vspace{-0.55cm}
\end{table}

A sequence of \textit{$14$ continuous frames} are used as input for each sensor data. During training, we use cross-entropy (two categories) as the loss function and apply an Adam optimizer \cite{kingma2017adam}.
For both Transformer models, the input embedding size ($D$), number of Transformer layers, and number of heads are set to $256, 8, 16$, respectively. The experiment results and execution time are shown in Table~\ref{tab:slipdetection}.
\subsubsection{Experiment Analysis}
From Table~\ref{tab:slipdetection}, we can see that the Transformer models can provide more accurate classification results. Also, when tested on the same dataset split as in \cite{li2018slip}, the Transformer models can achieve better results ($92.3$\% for TimeSformer and $90.0$\% for ViViT) than reported in \cite{li2018slip} ($88.0$\%) using both sensor inputs. One potential reason for the efficacy of Transformer models is that in this application, the final grasping outcomes may be inferred partially from the initial grasping status and Transformer models have the capacity of capturing these long-term temporal dependencies more effectively compared with recurrent networks \cite{zuo2021transformer}. Another potential reason is related to the architecture of the Transformer models: since each Transformer layer is stacked in a sequence, spatial and temporal information can be extracted simultaneously via self-attention mechanism, which does not hold for the CNN + LSTM models. 

In addition, it takes significantly less time for feed-forward computation of the trained networks during the robotic deployment. As shown in Table~\ref{tab:slipdetection}, using both vision \& tactile inputs from the same test dataset and selecting the same batch size, the execution time of both TimeSformer and ViViT models on the same machine (NVIDIA GeForce RTX 2070) is 
$2.46$ s ($25.6$\%) and $2.43$ s ($25.3$\%), compared to the CNN + LSTM model (ResNet18: $9.61$ s, VGG16: $21.39$ s). 
Therefore, it can be concluded that Transformer enables the robots to make decisions
within a much shorter time.

For the CNN+LSTM model, tactile-only significantly outperforms vision-only, as also reported in \cite{li2018slip}. The Transformer models perform similarly for each single sensor case while also showing better performance for the multi-sensor case. This indicates that multi-sensor input provides better cues for the slip detection.

These advantages highlighted above altogether motivate us to exploit Transformer models on safe fruit grasping.
\vspace{-0.4cm}
\subsection{Transformers for Safe Fruit Grasping}
\vspace{-0.1cm}
\begin{figure}[t]  
\centering
\includegraphics[width=\linewidth]{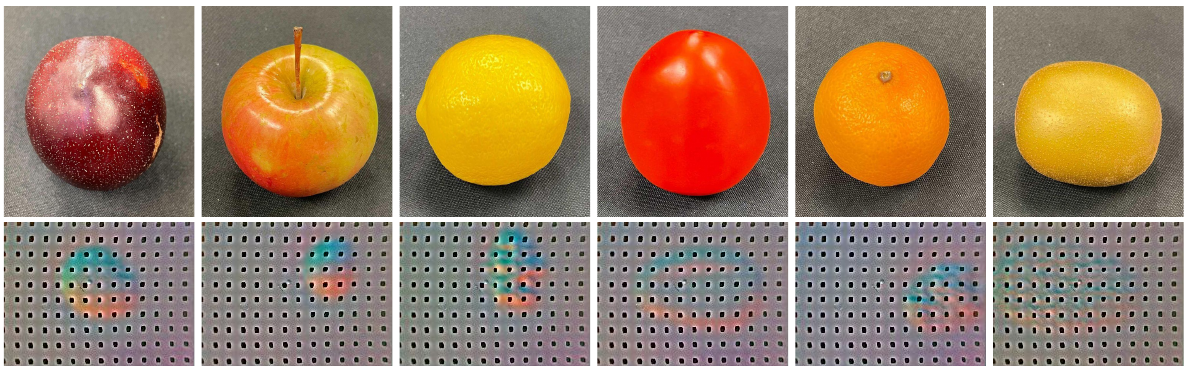}
\caption{Top row: fruits used in experiments. From the left to the right: plums, apples, lemons, tomatoes, oranges, and kiwifruits. 
Bottom row: the Gelsight images collected at the final frame during pinching for each fruit grasping. It can be seen that the fruit deformation sensed by Gelsight varies as they share different hardness and surface texture.}
\label{fig:fruit}
\vspace{-0.5cm}
\end{figure}
In this subsection, we examine our framework for grasping deformable fruits.
\subsubsection{Experiment Setup}
\label{see}
We collect our own dataset 
\footnote{\href{https://drive.google.com/file/d/144cLi-LkPZVHD\_JHfJSC8mYAk6gISCNI/view?usp=share\_link}{https://drive.google.com/file/d/144cLiLkPZVHD\_JHfJSC8mYAk6gISCNI/ view?usp=share\_link}} on fruit grasping involving six different types of fruits: plums, oranges, lemons, tomatoes, apples, and kiwifruits, as shown in Fig.~\ref{fig:fruit} (top row). We perform the fruit grasping with various grasping force thresholds (discussed in Sec\ref{sec:normal_force}) on them for $782$ times in total to train the models. For each type of fruit, due to the variations in fruit hardness and surface texture, the grasping force thresholds are different as we ensure a balanced training data distribution among the three grasping outcomes (i.e., the count numbers of three grasping outcomes are similar). For example, the sequence of force threshold used for apple (the hardest fruit) is sampled as integers from $4$ to $16$, and for orange (the softest fruit) it is from $4.0$ to $10.0$ with $0.5$ intervals (both have $13$ force thresholds). Since each fruit has a different range of force threshold, we include decimal values to guarantee that the numbers of each fruit samples in training data are close. 
Each fruit is clearly visible with a black backdrop relative to the camera frame. On the bottom row in Fig.~\ref{fig:fruit}, we show that the fruit deformation varies during pinching as they differ in hardness and surface texture. The data is collected by the RealSense camera and GelSight at $30$ Hz and with $640$$\times$$480$, $200$$\times$$150$ resolutions, respectively. The visual images are then resized to $160$$\times$$120$ resolution for computational efficiency. For both pinching and sliding actions, we use the \textit{first frame of every three continuous frames} for a total of $8$ frames (frame index: $1, 4, 7, 10, 13, 16, 19, 22$). 
For both Transformer models, the patch sizes are set as ($20,15$) for tactile data and ($16,12$) for visual data. The input embedding size ($D$), number of Transformer layers, and number of heads are set to be $256, 16, 8$, respectively. 

\subsubsection{Experimental Evaluation}
We aim to address the following questions in this experimental evaluation. 
\begin{itemize}
    \item Can our Transformer models outperform the CNN + LSTM models in terms of grasping outcome prediction for unseen fruits? 
    \item Is it plausible to deploy the trained frameworks for online fruit grasping applications?
    \item What patterns do the Transformer models learn from data? In other words, where do the Transformer models attend to?
\end{itemize}

\subsubsection{Grasping Outcome Prediction on Unseen Fruit}
\label{outcome}
We use a cross-validation technique, partitioning one type of fruit grasping data as a testing set and others as a training set, to compare the accuracy of grasping outcome prediction. After training, the Transformer models achieve $80.2$\% and $76.0$\% accuracy of grasping outcome prediction on the test dataset (kiwifruit) for TimeSformer and ViViT, respectively. For CNN + LSTM model with Resnet18 as the CNN architecture, it achieves $75.0$\% accuracy on the test dataset. \textit{building connection with the later training with banana.}

\subsubsection{Online Fruit Grasping Evaluation}
\label{grasping}
Then, the trained frameworks are deployed on a $7$-DOF KUKA LBR iiwa robot manipulator to estimate the safe grasping force via inference for both seen and unseen fruits. During inference, we sample the force thresholds as integers between $4$ and $16$, and for each sample, we adopt the same fused physical embedding obtained from performing two pre-defined explorative actions. The robot then grasps each fruit $50$ times. Table.~\ref{tab:onlinegrasping} shows the success rate and the average on-board computation time for one sample. 
It can be seen from Table.~\ref{tab:onlinegrasping} that the Transformer models outperform the CNN+LSTM model significantly, for both seen and unseen fruit grasping.

{We demonstrate the total computation time for each successful grasping. As we show in Table.~\ref{tab:onlinegrasping}, our framework with ViViT takes 0.29s for each force threshold sample. In total, we have 13 samples (all integers between 4 and 16), so the total on-board computation time is around 3.77s. We also allow 15 seconds for the endure time for the two explorative actions. As a result, for each successful grasping, it takes around 20 seconds in total, and this indicates our framework can accomplish 180 graspings per hour. It should be noted that since we only use CPU (11th Gen Intel(R) Core(TM) i9-11900K @ 3.50GHz) during online deployment, we believe our framework shows the potential for real-time industrial application, with further computational improvement or even GPU implementations.}



\begin{table}[t]
    \centering
    \begin{tabular}{|c|c|c|c|c|}
    \hline
        \multirow{3}{*}{\diagbox[width=11em,height=3\line]{Fruit}{Success Rate}{Model}} & \multirow{3}{*}{CNN + LSTM} & \multirow{3}{*}{TimeSformer} & \multirow{3}{*}{ViViT} \\ 
            &  & &  \\
            & ResNet18 & & \\
        \hline
    Plum    &  66\% & 88\% & 86\%\\
        \hline
    Orange  & 64\%  & 86\% & 90\% \\
        \hline
   Lemon    &  60\%  & 90\%  &  90\% \\
        \hline
    Tomato    & 74\%  & 86\% &  86\% \\
        \hline
    Apple    & 68\%  & 92\%  & 92\% \\
        \hline
    Kiwifruit (unseen)    & 52\%   & 74\% &  80\%\\
        \hline \hline
        Computation time& \multirow{2}{*}{0.52 s} & \multirow{2}{*}{0.39 s}  & \multirow{2}{*}{0.29 s}\\
        for one sample  & & & \\
        \hline
    \end{tabular}
    \caption{Experimental results of success rate for online fruit grasping (50 trials for each fruit) and the average computation time.}
    \label{tab:onlinegrasping}
    \vspace{-0.2cm}
\end{table}
\begin{figure}[t]  
\centering
\includegraphics[scale=0.45]{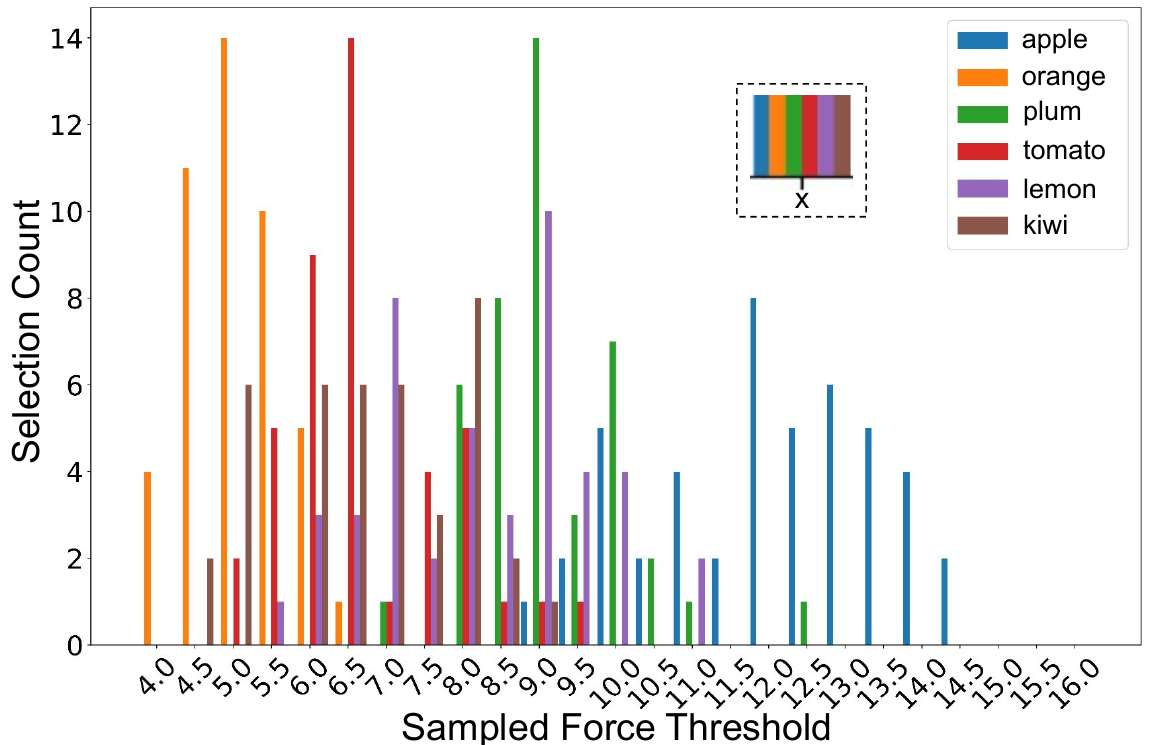}
\caption{The times each force threshold candidate is selected for safe grasping. The positional order of the color bars at each sample is shown in the dashed block.}
\label{fig:bar}
\vspace{-0.5cm}
\end{figure}
\begin{figure}[t]  
\centering
\includegraphics[width=\linewidth]{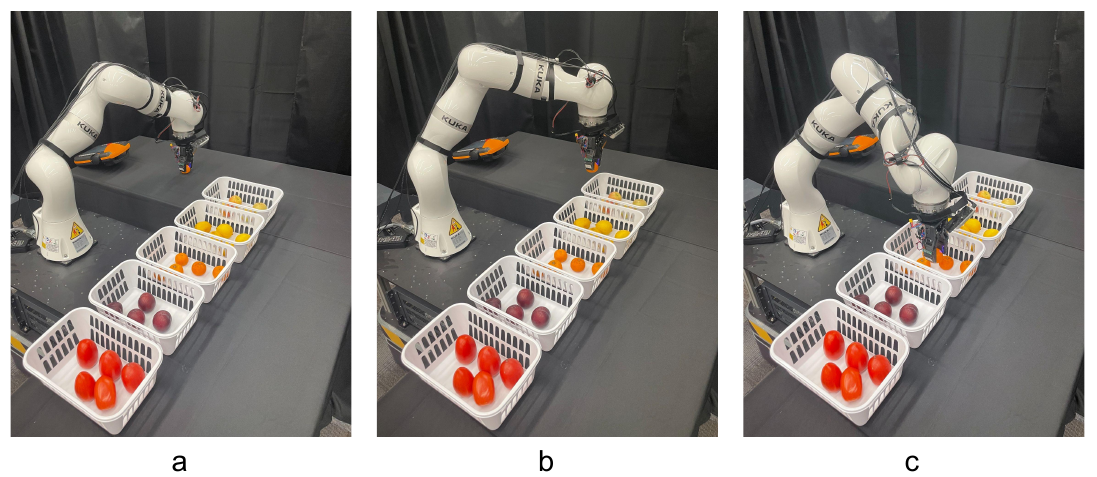}
\caption{Snapshots of fruit picking operation. (a): Grasp; (b): Move;
(c): Place. Other fruit experiments are shown in the attached video.}
\label{fig:fruit_picking}
\vspace{-0.2cm}
\end{figure}
It is noteworthy that in spite of the \textit{variations in grasping position} caused by manual fruit reloading, \textit{slight fruit spoilage} caused by squeezing, and \textit{unseen fruit type (kiwi)}, the framework is still able to select the safe grasping threshold for each grasp. 
This shows that the framework demonstrates some level of generalizability under the uncertainty of the local contact surface texture and fruit ripeness.
Besides, Fig.~\ref{fig:bar} shows the times each force threshold candidate is selected for the successful fruit grasping when using ViViT. Their values are proportional to the grasping force that should be exerted on the fruit. 
It is observed that the selected force thresholds for safe grasping of each fruit distribute over a finite range. Take orange as an example, force threshold $4.5$ is selected 11 times for safe grasping of \textit{softer oranges} and $6.0$ is also selected $5$ times for \textit{harder oranges}. 
This force threshold variation indicates our framework's adaptation to the fruit's inherent variability, which is infeasible by hard coding a fixed force threshold, even for the same fruit. 

We also test the Fruit Classification model on the seen fruits during online deployments (video can be found here \footnote{\href{https://www.youtube.com/watch?v=W7o8DsTivTk}{https://www.youtube.com/watch?v=W7o8DsTivTk}}), which enables the KUKA
LBR iiwa robot to place each fruit into separate bins using its built-in position-based waypoint tracking controller after successful grasp. For this, we pre-define five different waypoints for each fruit and when the robot reaches the desired waypoint, it would drop off the grasped fruit immediately.
In Fig.~\ref{fig:fruit_picking}, we show one case that the robot first grasps the orange from the table and then places it in the target bin using the proposed framework. It should be noted that the purpose of this operation is to illustrate that our framework can be potentially used for an integrated pick-and-place task. Therefore, our Fruit Classification model is not compared with other existing methods since it is not the focus of this work.
\subsubsection{Attention Analysis}
\label{attention}
\begin{figure}[tb!]  
\centering
\includegraphics[scale=0.65]{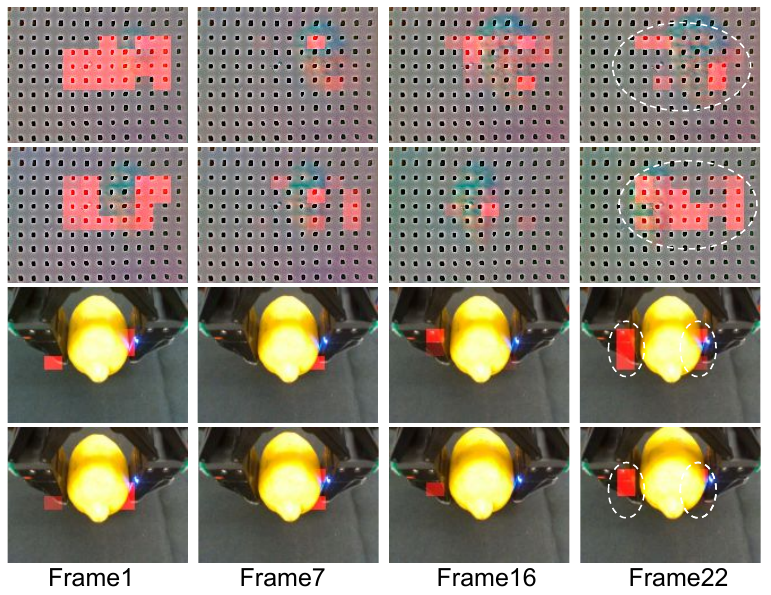}
\caption{Visualization of temporal attention from selected image patches at the final frame to their temporally preceding neighbours during a lemon grasping. We only show the results of four frames here. From top to bottom rows, the images are collected from pinching (tactile), sliding (tactile), pinching (visual) and sliding (visual), respectively.
The image at each frame is split into $10\times10$ patches, among which $24$ and $6$ patches are selected (denoted within the dashed ellipsoid in the final frame) to present the temporal attention flows for tactile and visual images, respectively. The brighter the patch color is, the more attention is paid to the patch from its temporal neighbor at the final frame.  
}
\label{fig:temporalAttention}
\vspace{-0.5cm}
\end{figure}

\begin{figure}[t]  
\centering
\includegraphics[scale=0.38]{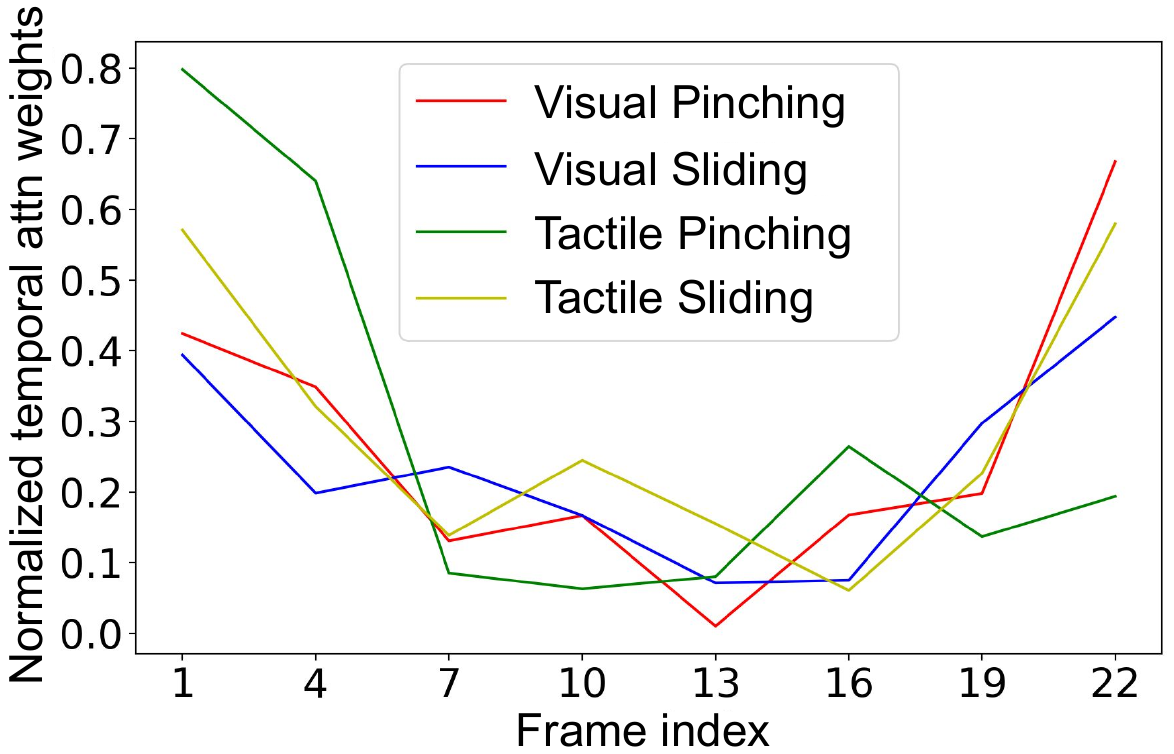}
\caption{The normalized temporal attention weights to all selected image patches at each frame from their corresponding temporal neighbors at the final frame.}
\label{fig:line}
\vspace{-0.3cm}
\end{figure}

\begin{figure}[htb!]  
\centering
\includegraphics[scale=0.66]{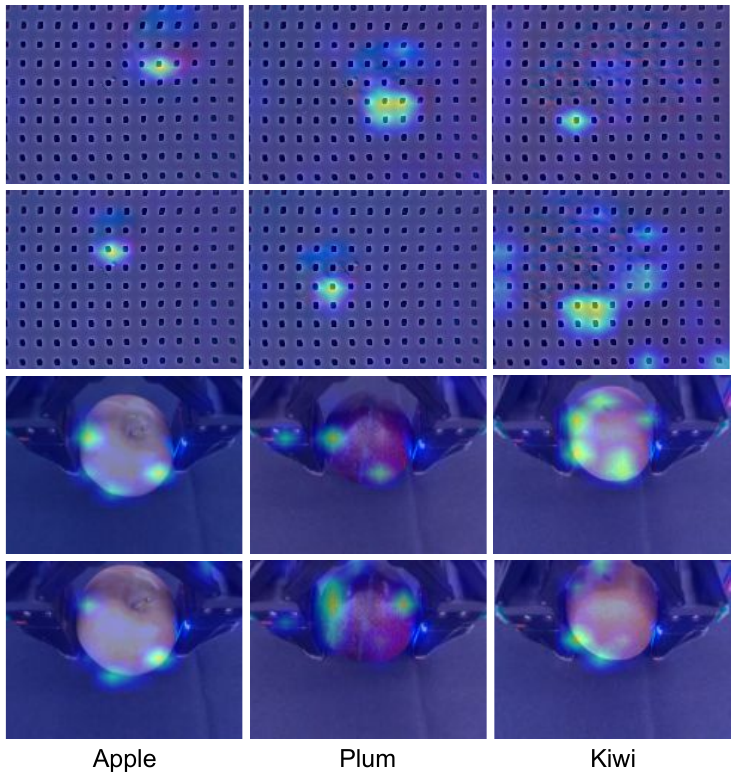}
\caption{Visualization of spatial attention from output token to the input image space at Frame $22$ during fruit grasping. From top to bottom, the images are collected from pinching (tactile), sliding (tactile), pinching (visual), and sliding (visual), respectively. The image brightness corresponds to spatial attention weights. 
It is clear that the model mostly attends to the \textit{local contact region} on tactile images and attends to the fruit surface near gripper's fingertips on visual images. 
It should be noted that kiwi is unseen during training.}
\label{fig:spatialAttention}
\vspace{-0.5cm}
\end{figure}
\vspace{-0.05cm}
One important component of our framework is the Transformer models learned entirely from data. Therefore, we now examine what pattern has our models learned qualitatively.
Take TimeSformer as an example, we use the Attention Rollout method \cite {abnar2020quantifying} to visualize the learned temporal attentions across vision \& tactile image sequences on several selected image patches, as shown in Fig.~\ref{fig:temporalAttention}. It can be seen that the image patches at the final frame do not only attend to themselves, but also their temporal neighbors at  preceding frames (red color brightness denotes the attention weights). In addition, Fig.~\ref{fig:line} shows the normalized temporal attention weights of all selected image patches. An intriguing observation is that the image patches at the first two frames, when the gripper initially touches the objects, share larger attention weights compared with succeeding intermediate frames. Our conjecture of this observation is due to the fact that as the initial and ending contact information is more inferable to the physical status of the manipulated objects, as well as to the grasping outcome. On the contrary, the gradient flow in recurrent networks, even for LSTM architecture, can gradually lose the information on the previous inputs, especially of the first few inputs, resulting in the difficulty of capturing long-term temporal dependencies \cite{Hochreiter01Gradient}. However, Transformer is able to mitigate this problem as demonstrated. 

Furthermore, we show the spatial attention at the final frame for apple, plum, (seen during training) and kiwifruit (unseen during training) grasping in Fig.~\ref{fig:spatialAttention}. For tactile images, the TimeSformer model mostly attends to the local contact region, and for visual images, it attends to the fruit surface near gripper's fingertips. Therefore, the Transformer models can incorporate more contact information for the grasping task.

We highlight that the interpretability of attention mechanisms may provide an alternative way of analyzing how deep learning methods understand the object's physical deformation properties captured by tactile and visual sensors during contact-rich tasks. 

\begin{figure}[t]  
\centering
\includegraphics[width=\linewidth]{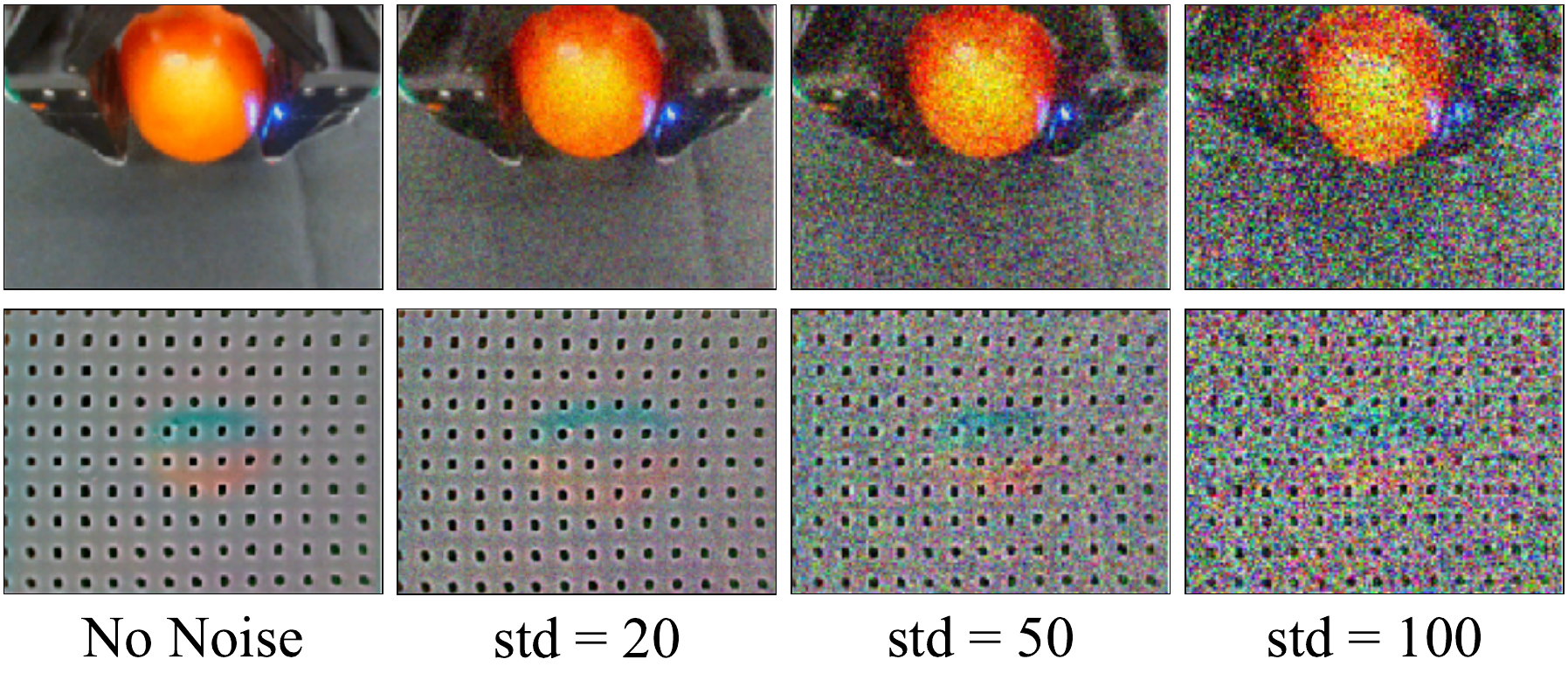}
\caption{Top row: Visual images for tomato with different noises.  
Bottom row: Gelsight images for tomatoes with different noises. From left to right, the Gaussian noises added on images are no noise, $\sigma = 20, 50, 100$, respectively.}
\label{fig:noise}
\vspace{-0.2cm}
\end{figure}

\begin{figure}[t]  
\centering
\includegraphics[width=\linewidth]{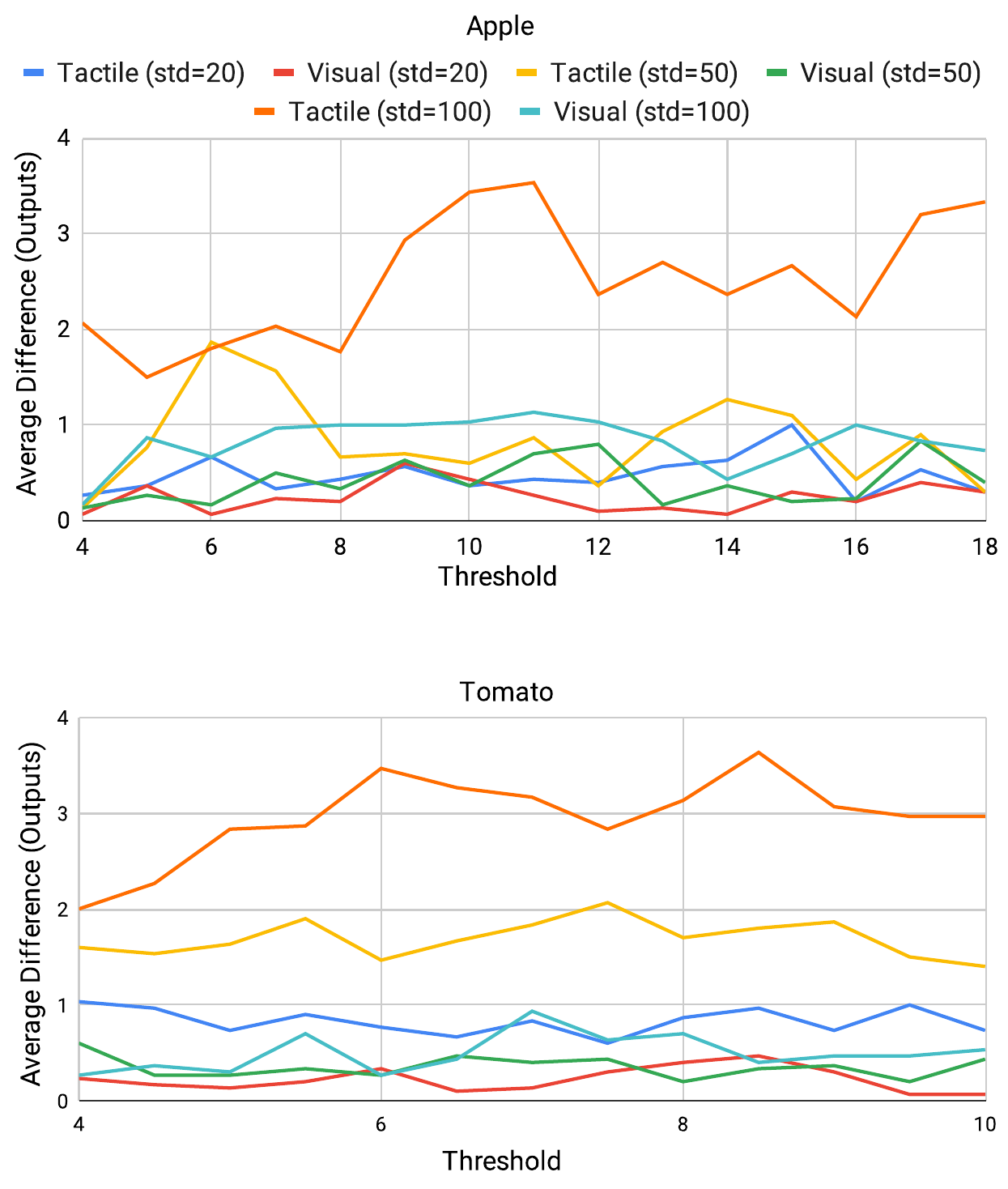}
\caption{These two plots show the relationship between Sample Force Thresholds (x-axis) and Average outcomes Difference (y-axis) with noise and the outcomes without noise for apples and tomato. The lines with different colors represent different levels of Gaussian noises.}
\label{fig:difference}
\vspace{-0.5cm}
\end{figure}

\begin{figure}[t]  
\centering
\includegraphics[width=\linewidth]{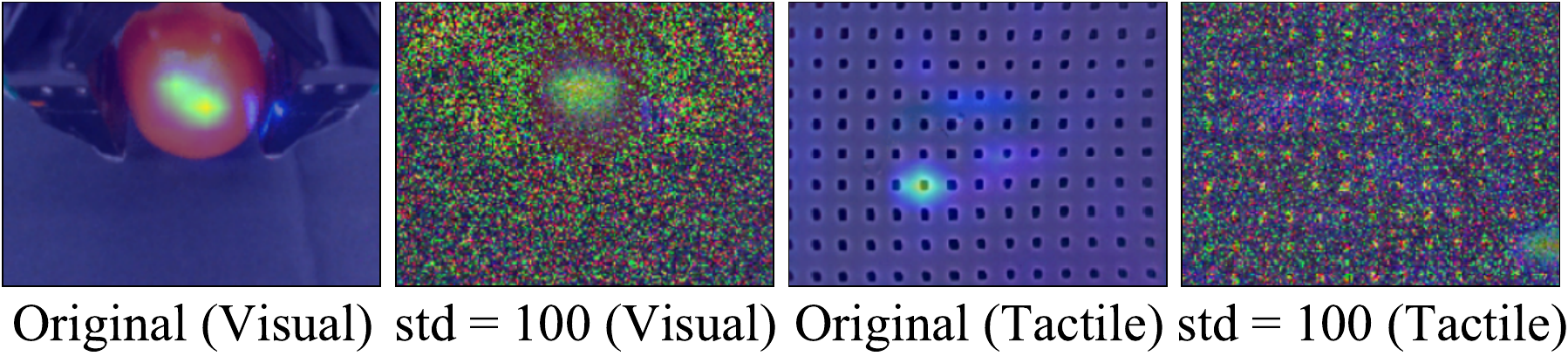}
\caption{This figure shows the comparison of spatial attention results for the tomato in both visual and tactile images between zero noise and $\sigma = 100$.}
\label{fig:noise_attention}
\end{figure}

{\section{Additional Experiments and Analysis}
\subsection{Sensitivity Analysis of Visual and Tactile Feedback}
In this section, we evaluate the sensitivity of our framework in visual and tactile images with different data qualities.
\subsubsection{Experiment Setup}
\label{mulset}
We employ our proprietary fruit-grasping dataset for sensitivity analysis, aiming to evaluate our framework's performance across varying image qualities. We add different noises to the images in our training set and evaluated the performance with the trained model. We assume Gaussian noise, mimicking natural variations and imperfections in real images. Mathematically, Gaussian noise added on an image is shown as $\mathcal{N}(0, \sigma)$, where \(\mathcal{N}(0, \sigma)\) represents the Gaussian distribution with a mean of $0$ and a standard deviation of \(\sigma\), which is a pixel value in range $[0, 255]$. \\
Gaussian with different standard deviations $(\sigma = 20, 50, 100)$ are added to the images, as shown in Fig.~\ref{fig:noise}. We use apples (hard fruit) and tomatoes (soft fruit) fromin our training set for analysis. The force thresholds used for apples are integers from $4$ to $18$, while tomatoes employed force thresholds ranging from $4.0$ to $11.0$ with $0.5$ intervals. For each input, our framework predicted $3$ values, such as $[14.7, 3.7, -27.9]$ for each label of $[0, 1, 2]$ which represented [slipping, safe grasping, and damage], selecting the label with the highest value. To assess the impact of noise and thresholds, $10$ model predictions are conducted for each level of noise and threshold, calculating the absolute value of the average difference between outcomes with and without noise. For example, an outcome without noise $[14.7, 3.7, -27.9]$, and an average outcome with noise $[14.7, 3.8, -28.0]$, resulted in differences of $[0, 0.1, 0.1]$. The average difference is then calculated as $0.067$, representing the final result. This procedure is repeated for tactile and visual images of both apples and tomatoes, encompassing $15$ different thresholds and $3$ noise levels.}
{\subsubsection{Experiment Evaluation}
\label{mulmodal}
Upon conducting experiments with both tactile images and visual images of apples and tomatoes, 
we present the results in Fig.~\ref{fig:difference}. Visual noise analysis reveals the largest difference in outcomes at the highest noise level $(\sigma = 100)$. However, when considering the average outcomes in Sec. \ref{mulset}, this discrepancy represents only $6.5\%$ of the outcomes. Hence, visual noise does not significantly impact the model's prediction performance. Conversely, tactile image noise exhibits a more pronounced difference. Substantial tactile noise $(\sigma = 100)$ can lead to a difference of approximately $3.5$, accounting for $20\%$ of the outcomes.
The results in both apples and tomatoes validate the superior robustness of our framework to visual images than tactile images
}
{\subsubsection{Attention Analysis}
In addition, we conducted attention analysis on images with varying levels of noise using the same method as described in Sec. \ref{attention}. Our focus in this and subsequent subsections is to evaluate spatial attention.
When visualizing the spatial attention of apple with different noise levels (see Fig.~\ref{fig:noise_attention}), we observe that the TimeSformer model primarily attended to localized contact regions in visual images, even in the presence of high noise levels ($\sigma = 100$). However, under the same noise conditions, the model struggled to effectively attend to the fruit surface in tactile images. These findings further solidify our conclusion on the superior robustness of our framework to visual images than tactile images.}

\begin{figure}[t]  
\centering
\includegraphics[width=\linewidth]{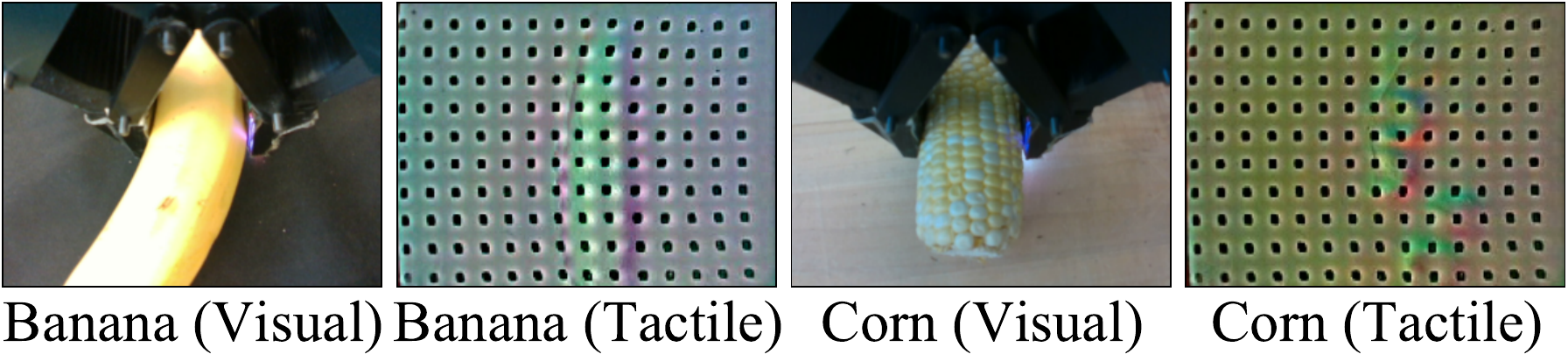}
\caption{This figure shows the visual images and Gelsight images for corn and banana.}
\label{fig:new_gripper}
\vspace{-0.5cm}
\end{figure}

\vspace{-0.5cm}
{\subsection{Fruit Grasping Evaluation for Unseen Irregular Objects}
In this section, we evaluate our framework with unseen irregular objects (i.e., corn and banana) during online experiments, the same as Sec. \ref{see}. Unlike kiwi, corn and bananas are new objects with irregular shapes and different contact surface textures, as shown in Fig.~\ref{fig:new_gripper}.}
{Note that, we use a new gripper in this section for this experiment, which has almost exactly the same mechanical properties as the previous one.}
{\subsubsection{Experiment Evaluation}
\label{corn}
We deploy our model trained from five different round fruits shown in Sec. \ref{see} on corn and banana to estimate the safe grasping. Similar to the experiments for other fruits, we sample the force threshold as integers between $4$ and $16$. The robot then grasps each fruit $50$ times. Table.~\ref{tab:corn} shows the success rate.}

{Analyzing the results depicted in Table.~\ref{tab:corn}, we observe that our model achieves a high accuracy of $72\%$ for the corn grasping, where the visual and tactile images of the corn have different patterns from those of the round fruits in our training set. This result
highlights the robustness of our model in dealing with varying visual images. However, the accuracy for bananas is comparatively lower at $38\%$, indicating limited generalizability when encountering highly varying tactile images. Fig.~\ref{fig:new_gripper} illustrates the distinctive contact surface textures of a banana compared to the circular contact surface textures typically found in fruits with a round geometry. While the contact surface textures of corn are also irregular, our conjecture is that it consists of multiple smaller circles that bear resemblance to round objects, enabling our model to perform well in corn-grasping tasks.}
\begin{table}[t]
    \centering
    \begin{tabular}{|c|c|c|c|c|}
    \hline
        \multirow{3}{*}{\diagbox[width=11em,height=3\line]{Fruit}{Success Rate}{Model}} & \multirow{3}{*}{CNN + LSTM} & \multirow{3}{*}{TimeSformer} & \multirow{3}{*}{ViViT} \\ 
            &  & &  \\
            & ResNet18 & & \\
        \hline
    Corn    &  46\% & 68\% & 72\%\\
        \hline
    Banana  & 18\%  & 38\% & 38\% \\
        \hline
    \end{tabular}
    \caption{Experimental results of success rate for online fruit grasping ($50$ trials for each fruit).}
    \label{tab:corn}
    \vspace{-0.5cm}
\end{table}
\subsubsection{{Attention Analysis}}

\begin{wrapfigure}[9]{r}{0.5\linewidth}
\centering
\includegraphics[width=\linewidth]{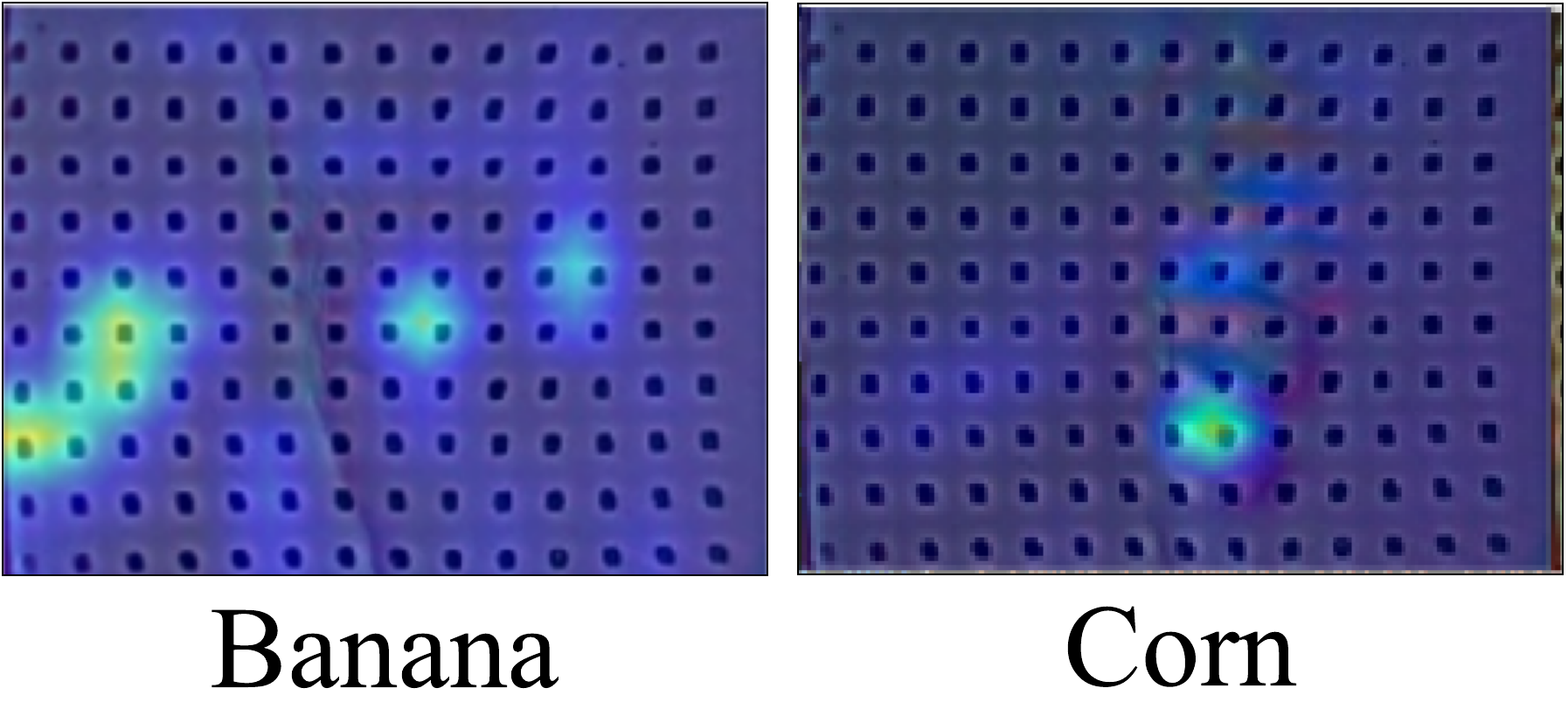}
\caption{This figure visualizes the spatial attention for tactile images of a corn and a banana, respectively.}
\label{fig:new_gripper_attention}
\end{wrapfigure}

We visualize the spatial attention of
corn and bananas using the same method as Sec. \ref{attention}. Fig.~\ref{fig:new_gripper_attention} illustrates 
our model can attend to the contact surface in tactile images of corn, but it's not doable in bananas. 

Again, we attribute the reason to the fact that the contact surface of corn is composed of small circles, which are similar to the training objects. 
However, our model fails to attend to either the banana or the corn in visual images due to their disparate camera views. 

\vspace{-0.4cm}
\subsection{Fine-tuning the Pre-trained Model for an Object with New Geometries}
Since our model cannot generalize to the banana because of the significant shape difference, we collect a banana dataset\footnote{\href{https://drive.google.com/file/d/1ZVHpKwplXh-sKuNIiLo_m5lEbJmXVE76/view?usp=sharing}{https://drive.google.com/file/}} and train a new model to validate that our framework is applicable to bananas. Then, we examine our framework for grasping bananas after fine-tuning  a pre-trained model on the banana dataset.
\subsubsection{Experiment Setup}
We use the same method as Sec. \ref{see} to collect a new dataset on banana grasping. We perform the fruit grasping with various grasping force thresholds on them $64$ times. 
The sequence of force thresholds uses for bananas ranges from $3.0$ to $6.5$ with $0.5$ intervals. 
\subsubsection{Training}
When acquiring a new dataset, it is common to result in smaller dataset sizes compared to the original dataset. For example, the size of our banana dataset is only one-tenth of that of the original dataset. Owing to the scarcity of data, training the model solely on the new dataset often fails to yield satisfactory performance. In this section, we aim to leverage the model trained on the original dataset described in Sec. \ref{outcome} as a pre-trained model to solve this challenge.

During the training process, we split the dataset into a training set and a validation set with a ratio of $7:1$.

When training solely with the banana dataset, each epoch requires approximately $23$ seconds.
When using the pre-trained model, the time per epoch is reduced to $8$ seconds.
This represents a significant reduction in training time, leading to a substantial improvement in training efficiency.

In terms of training results, employing the pre-trained model results in a training accuracy of $75$\%, surpassing the accuracy achieves without the pre-trained model by $10\%$. Moreover, training without the pre-trained model exhibits large oscillations in both validation loss and accuracy due to the limited size and single type of the banana dataset. 
In thia case, employing our pre-trained model, which is trained on a larger dataset, mitigates these oscillations and enhances the model's performance.
\subsubsection{Experiment Evaluation}
We deploy our model trained from the banana dataset to a real banana grasping experiment to estimate the online grasping results. Similar to the setting in the Sec. \ref{grasping}, we sample the force thresholds as integers between $3$ and $12$. For each sample, we adopt the same fused physical embedding obtained from performing two pre-defined explorative actions. In Table.~\ref{tab:pre-trained}, we present the results obtained from three different scenarios: the original model, which is not trained on the banana dataset; the model trained solely on the banana dataset; and the model trained on the banana dataset using the pre-trained model.
\begin{table}[t]
    \centering
    \begin{tabular}{|c|c|c|c|c|}
    \hline
        \multirow{3}{*}{\diagbox[width=12em,height=3\line]{Fruit}{Success Rate}{Model}} & \multirow{3}{*}{CNN + LSTM} & \multirow{3}{*}{TimeSformer} & \multirow{3}{*}{ViViT} \\ 
            &  & &  \\
            & ResNet18 & & \\
        \hline
    Banana (w/o training)    &  18\% & 38\% & 38\%\\
        \hline
    Banana (w/o pre-trained)  & 48\%  & 76\% & 76\% \\
        \hline
    Banana (with pre-trained)  & 52\%  & 80\% & 84\% \\
        \hline
    \end{tabular}
    \caption{Experimental results of success rate for online banana grasping with different models ($50$ trials for each model).}
    \label{tab:pre-trained}
    \vspace{-0.2cm}
\end{table}

{After training on the banana set, the model reveals a significant performance improvement. This demonstrates the effective learning capabilities of our model, even when dealing with fruits of complex and irregular shapes, leading to enhanced grasping performance. Also, the performance gap becomes more pronounced when handling intricate, irregular objects such as the banana. With the use of the pre-trained model, our framework achieves a high success rate of up to $84$\%, while the baseline model reaches only $52$\%.}

\begin{figure}[t]  
\centering
\includegraphics[width=\linewidth]{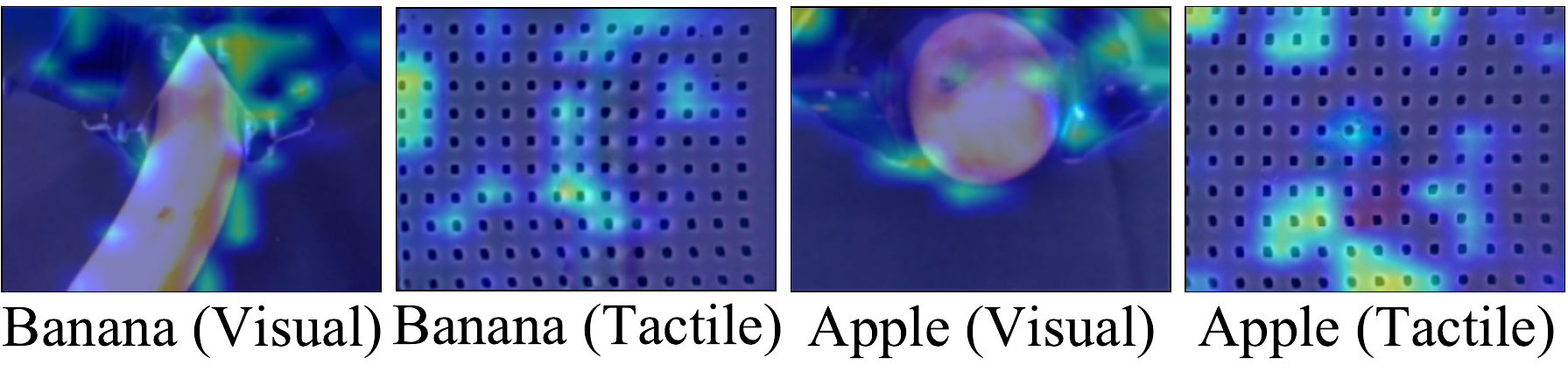}
\caption{This figure visualizes the spatial attention of banana and apple after training with the pre-trained model.}
\label{fig:pre-trained}
\vspace{-0.3cm}
\end{figure}

{Furthermore, the advantages of employing the pre-trained model successfully extend to the online grasping experiment. When using the pre-trained model, our framework achieves a higher grasping success rate of $84$\%, compared to $76$\% without employing the pre-trained model. It highlights the versatility of our model, as it can not only be used in safe grasping for round fruits but also serve as a pre-trained model for training on new datasets, even when the new fruit exhibits a completely different and irregular shape.
\subsubsection{Attention Analysis}
Same as Sec. \ref{attention}, we examine the performance of spatial attention after training with the banana set. As shown in Fig.~\ref{fig:pre-trained}, the TimeSformer model can mostly attend to the local contact region of visual images and fruit surface of tactile images for both banana and apple (in the original dataset) after training with the banana set, although they have very different shapes and contact surface textures. Therefore, the Transformer models can incorporate more contact information for the grasping task, even with fruits with irregular shapes.}

\section{Conclusions and Discussions}
\linespread{1.0}
Our experiments demonstrate that the Transformer models can enable robotic grasping tasks in both the object classification and robot control domain. The results indicate that they outperform traditional models, such as CNN+LSTM, for classification tasks like slip detection and grasping outcome prediction. In addition, our Transformer-based grasping framework is able to select the grasping strength to safely grasp fruits with varying hardness and surface texture. We also visualize the attention flows of the Transformer models, which can potentially explain their effectiveness and efficacy. {With the attention analysis, we find that our model has greater robustness with various data qualities in visual images than that in tactile images. Furthermore, our model can effectively learn from fruits with complex and irregular shapes and serve as a pre-trained model for training on new datasets.} However, it is worth noting that the Transformer models are still model-free methods relying on the learned attention from rich data. Performance could be expected to be improved by incorporating model-based methods, such as physical contact models, as future work. Also, improving grasping task robustness and generalization via adversarially regularized policy learning \cite{zhao2021adversarially} would be another direction to explore. On the other hand, considering another common scenario of grasping objects in a cluttered scene, we can integrate our framework with a high-level task planner to decide the collision-free grasping sequence \cite{Wang2022Cluttered} \cite{Sundermeyer2021GraspNet}.
{
\bibliographystyle{./IEEEtran}
\bibliography{IEEEcitation}
}
\end{document}